\documentclass{article}
\usepackage{ijcai24}

\usepackage{times}
\usepackage{soul}
\usepackage{url}
\usepackage[hidelinks]{hyperref}
\usepackage[utf8]{inputenc}
\usepackage[small]{caption}
\usepackage{graphicx}
\usepackage{amsmath}
\usepackage{amssymb}
\usepackage{amsthm}
\usepackage{booktabs}
\usepackage{xcolor}
\usepackage[linesnumbered,ruled,vlined]{algorithm2e} 

\usepackage[switch]{lineno}
\usepackage{multirow} 

\usepackage{graphicx}

\newcommand{\Checkmark}{\checkmark} 

\title{LERENet: Eliminating Intra-class Differences for Metal Surface Defect Few-shot Semantic Segmentation}

\author{
Hanze Ding$^1$
\and
Zhangkai Wu$^2$\and
Jiyan Zhang$^{1}$\And
Ming Ping$^1$\\
Yanfang Liu$^1$\\
\affiliations
$^1$Longyan University\\
$^2$University of Technology Sydney\\
\emails
Ding\_H2@126.com,
berenwu1938@gmail.com,
\{lydeson,pengming2013\}@126.com,
liuyanfang003@163.com
}

\begin{document}

\maketitle

\begin{abstract}
Few-shot segmentation models excel in metal defect detection due to their rapid generalization ability to new classes and pixel-level segmentation, rendering them ideal for addressing data scarcity issues and achieving refined object delineation in industrial applications.
Existing works neglect the \textit{Intra-Class Differences}, inherent in metal surface defect data, which hinders the model from learning sufficient knowledge from the support set to guide the query set segmentation. 
Specifically, it can be categorized into two types: the \textit{Semantic Difference} induced by internal factors in metal samples and the \textit{Distortion Difference} caused by external factors of surroundings. 
To address these differences, we introduce a \textbf{L}ocal d\textbf{E}scriptor based \textbf{R}easoning and \textbf{E}xcitation \textbf{Net}work (\textbf{LERENet}) to learn the two-view guidance, i.e., local and global information from the graph and feature space, and fuse them to segment precisely. 
Since the relation structure of local features embedded in graph space will help to eliminate \textit{Semantic Difference}, we employ Multi-Prototype Reasoning (MPR) module, extracting local descriptors based prototypes and analyzing local-view feature relevance in support-query pairs. 
Besides, due to the global information that will assist in countering the \textit{Distortion Difference} in observations, we utilize Multi-Prototype Excitation (MPE) module to capture the global-view relations in support-query pairs. 
Finally, we employ an Information Fusion Module (IFM) to fuse learned prototypes in local and global views to generate pixel-level masks. 
Our comprehensive experiments on defect datasets demonstrate that it outperforms existing benchmarks, establishing a new state-of-the-art.

\end{abstract}

\begin{figure}[!t]
\centering
\includegraphics[width=1\linewidth]{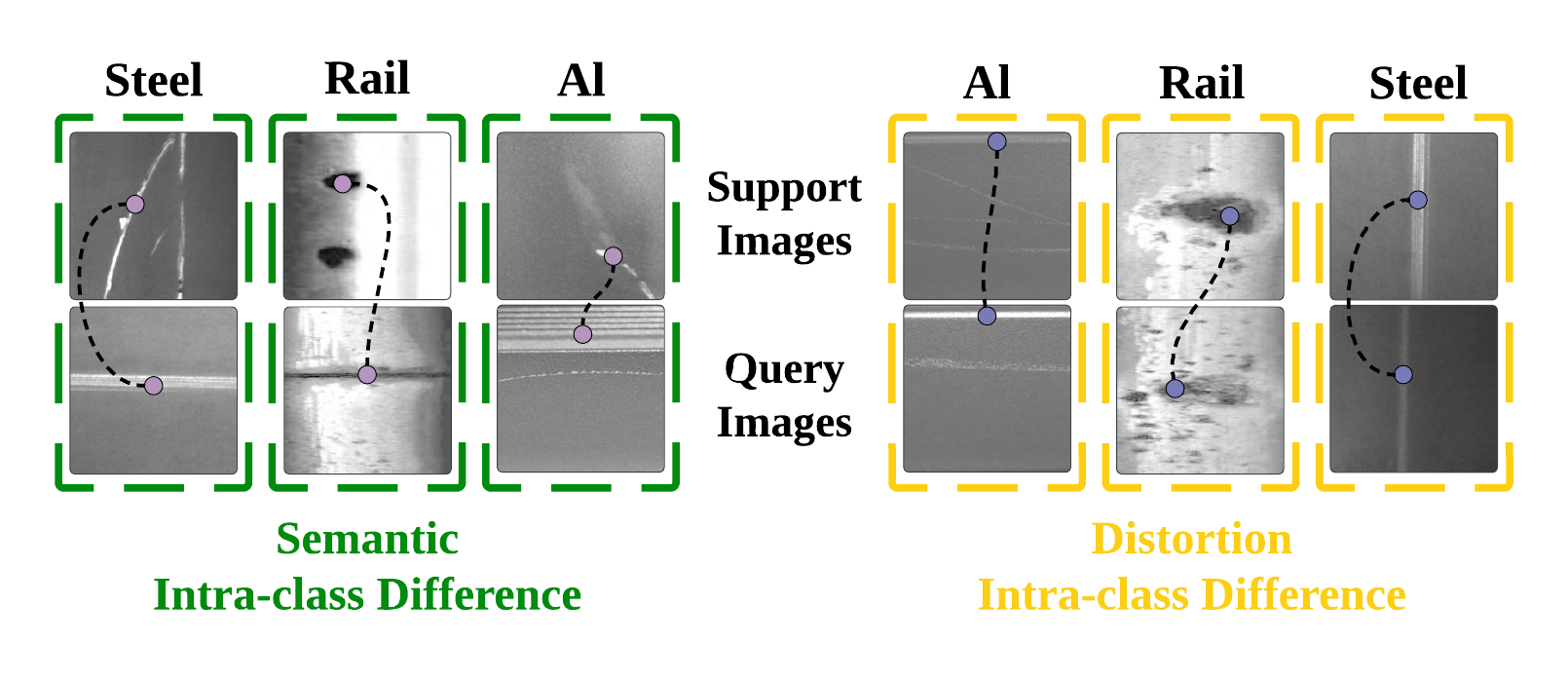} 
 \caption{Two categories of intra-class differences are observed in metal surface defect data. We characterize the semantic intra-class difference in $3$ support-and-query pairs, where defects, vary in fine-grained categories within the same categories. For instance, defects within the same class, such as Steel, Rail, or Al (aluminum), may appear differently under different manufacturing processes, lighting conditions, or if infected by various noises. We further identify the distortion intra-class difference in $3$ support-and-query pairs. Here, defects are induced by lens distortions or the perspective from which the image is taken, such as changes in shape, scale, and orientation in the defect instance.}
\label{F:difference}	
\end{figure}

\section{Introduction}
Few-Shot Segmentation (FSS) models demonstrate their potential in metal surface defect detection tasks. Recur to the few-shot learning ability \cite{finn2017model,snell2017prototypical,wang2019panet}, it can learn unseen concepts from a few annotated examples, tackling the data scarcity problem arising from the costly pixel-level annotations and rarity of the defect in the real industrial scenario. Beyond their rapid generalization capabilities, FSS models, enhanced by semantic segmentation modules \cite{yang2021mining,tian2020differentiable,zhang2022generalizable}, can accurately capture the location and structure of defects. The capability to make dense predictions on images is practical for industrial applications, overcoming the vague location information induced by classification based approaches and the inflexible boundaries caused by object detection methods where standard bounding boxes cannot adapt to the varied shapes of defects, such as patches and scratches \cite{DBLP:journals/tim/HeSMY20,luo2020automated}.

Aside from the single metal surface defect segmentation tasks \cite{pasadas2019defect,xie2016novel}, FSS based models demonstrate their potential in generic surface defect segmentation tasks \cite{bao2021triplet,yu2022selective}. However, these models disregard the intra-class differences inherent in metal defect samples. Consequently, the guidance information learned from the support set is insufficient in segmenting the samples within the query set. Specifically, we generalize it \cite{yang2023mianet} into two specific categories inherent in metal defect data: firstly, semantic intra-class differences stemming from the internal physical properties, where defects, although originating from identical metal types exhibit variations in their fine-grained classes. Secondly, distortion intra-class differences arise due to external factors, predominantly due to perspective distortion. For a more intuitive understanding of the two differences, refer to Figure \ref{F:difference}.

\begin{figure}[!t]
\centering
\includegraphics[width=1\linewidth]{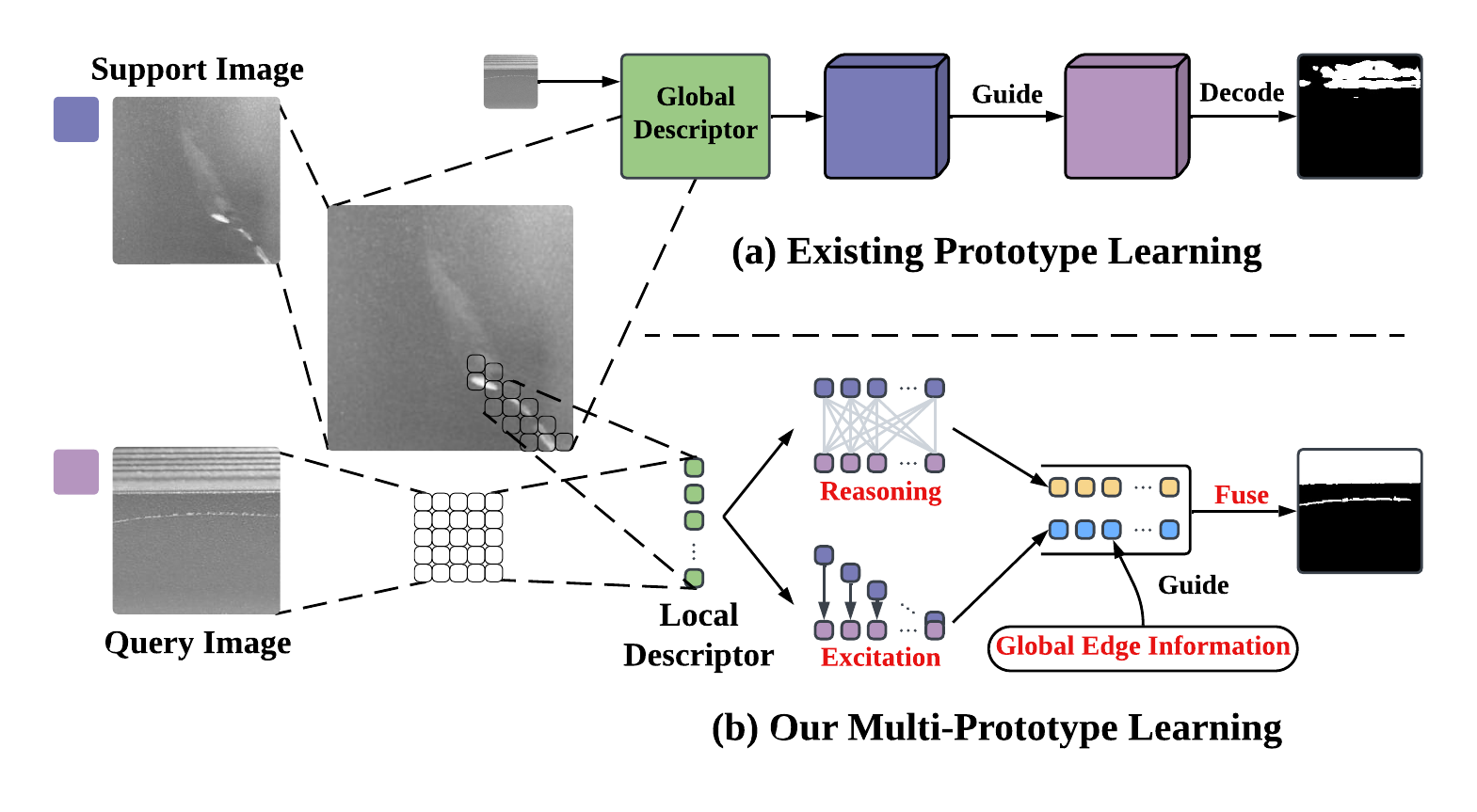} 
\caption{Comparison of traditional approaches and our LERENet. Firstly, traditional models extract features from image based prototypes. In contrast, LERENet employs local descriptor based multi-prototypes to represent more implicit local relations. Secondly, our model generates features in local (by Reasoning operation) and global views (by Excitation operation and Global Edge Infomation operation), respectively. After acquiring the local-view graph space features (represented by the yellow square) and the global-view features (represented by blue squares), the two types of differences are addressed separately.}
\label{F:motivation}	
\end{figure}

To address this, we propose LERENet, which consists of Multi-Prototype Reasoning (MPR) module, Multi-Prototype Excitation (MPE) module, and an Information Fusion Module (IFM) to produce sufficient guidance from multi-prototype-based support-and-query pairs to generate precise query masks. Different from traditional methods, we utilize deep local descriptors \cite{li2019revisiting} for feature embeddings, substituting the image-level features into local descriptors-level features to provide more flexible manipulation (the more comparison can be found in Figure \ref{F:motivation}). 

To diminish the semantic intra-class differences, which require the enhanced perception of local-view information within the same classes and the Graph Convolutional Network (GCN) excels in discerning correlation structures \cite{MESGARAN2024109960,liu2023few}. Motivated by that, we designed the MPR module. Specifically, we model the semantic correlations between support-query prototypes within a graph space, which are generated by local descriptors and the relevance of consistent defect features in support-query pairs can be achieved through reasoning among graph nodes.

We utilize the MPE module to tackle the distortion intra-class differences. In such scenarios, differences can be effectively mitigated by comprehensive global-view features, as distortion-induced perceptual blurring still contains the semantic information distinguished intuitively. For example, the ability to recognize a person from afar simply by observing their eyes demonstrates that critical global semantic details remain identifiable. Inspired by this, we activate relevant defect features directly in the feature space and extract global-view features by global edge information operations.

Finally, we fuse the local and global features from the MPR and MPE modules in the IFM. This fusion is employed to enrich the guidance from the graph and feature space, thereby leading to more accurate defect segmentation results. The more comparison and intuitive explanation of the LERENet framework can be found in Figure \ref{F:motivation}.

Our contributions can be summarized as follows:
\begin{enumerate}
\item[1)] We define two intra-class differences in FSS based metal defect tasks. Namely, semantic intra-class differences arise from the internal property within metal data and distortion intra-class differences arise from external factors in data collection.
\item[2)] We propose the MPR and MPE modules to generate multi-prototypes  
 based guidance information and employ IFM to do pixel-level segmentation. Experimental results show that our features fused by graph and feature space can alleviate the two differences effectively. 
 \item[3)] Numerous experiments have shown that the mIoU and FB-IoU performances of LERENet exceed those of the popular metal surface defect and amateur FSS networks, reaching state-of-the-art levels.
\end{enumerate}

\section{Few-shot Segmentation Learning in Metal Defect Detection}
\textbf{Metal Surface Defect Segmentation}. 
Metal surface defect segmentation is a crucial task in quality control during production and manufacturing stages in industrial scenarios, aiming to categorize each pixel in metal surface images into predefined semantic classes. Recent advances in metal semantic segmentation utilize multi-scale attention feature fusion modules for enhanced defect feature extraction \cite{zhang2023semantic} and integrate object detection with semantic segmentation to address diverse atypical defects \cite{sharma2022amalgamation}. However, these methods excessively depend on intra-class information, leading to suboptimal adaptability in scenarios with significant intra-class variations, often attributed to the uneven textural features of metal surfaces.

\textbf{Few-Shot Segmentation}. 
FSS is an extension of few-shot learning, to do pixel-wise prediction on unseen classes with limited samples available. OSLSM \cite{shaban2017one} first proposed the concept of few-shot semantic segmentation, in which the dual branch network is employed in meta-learning frameworks. FIB \cite{HU2024109993} applies the information bottleneck theory to few-shot semantic segmentation, which addresses the feature undermining problem for the target class. In recent years, the graph-based FSS algorithm has been proposed for metal surface defect segmentation \cite{bao2021triplet} tasks, which effectively reasons the potential relationship between the defect of the support-query pair. 

\textbf{Prototype-based Learning.} 
Prototype-based learning is a type of metric-based learning \cite{li2019distribution,zhang2023prototype}. It aims to learn a prototype for representation. Then perform the required downstream tasks by calculating the distance among the prototype. In recent years, query-guided prototype learning \cite{10.1145/3555314} points out and addresses the disadvantages of conventional prototype learning. TPSN \cite{WANG2022108326} utilizes a multi-prototype approach to address the issue that a single prototype cannot accurately describe a category. To address the challenges of conventional prototype learning, a local descriptor-based multi-prototype learning is proposed \cite{huang2021local}.

\begin{figure*}[!t]
\centering
\includegraphics[width=0.95\linewidth]{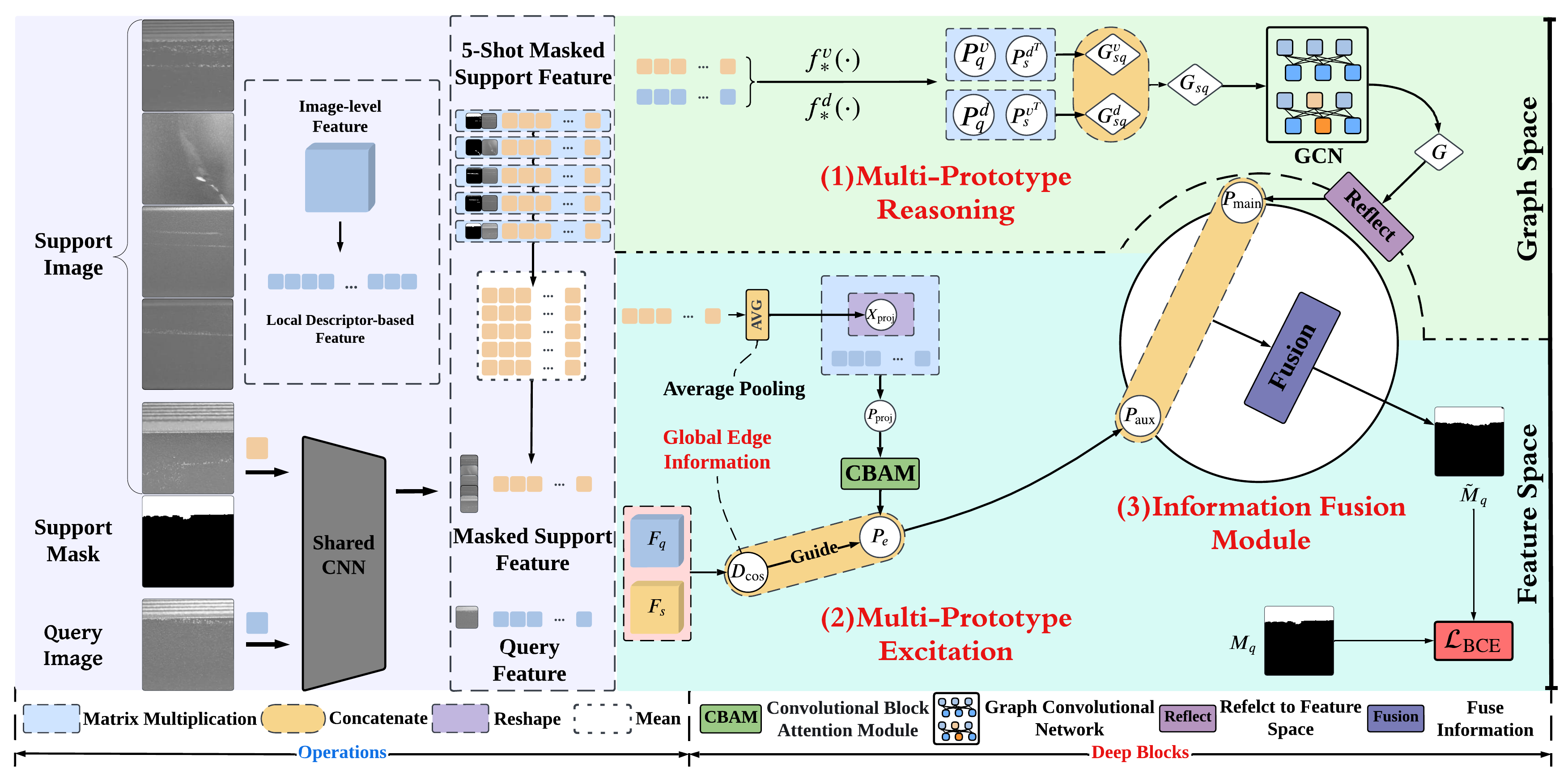}
\caption{LERENet for 5-shot segmentation. (1) represents the process of Multi-Prototype Reasoning (MPR). (2) denotes the Multi-Prototype Excitation (MPE). Given $\boldsymbol{P}_\mathrm{main}$ and $\boldsymbol{P}_\mathrm{aux}$ from the above steps, we will get the prediction $\tilde{\boldsymbol{M}}_{q}$ by (3) Information Fusion Module (IFM). Finally, we utilize BCE loss to train our model.}
\label{F:frame}
\end{figure*}

\section{Preliminary}
\textbf{Problem Definition.}
The model is trained on the training dataset $\boldsymbol{D}_{\mathrm{train}}$ and evaluated on the testing dataset $\boldsymbol{D}_{\mathrm{test}}$, with the metal classes sets $\boldsymbol{C}_{\mathrm{train}}$ and $\boldsymbol{C}_{\mathrm{test}}$ being disjoint.
Different from traditional semantic segmentation models \cite{ZHANG2024110018,10341333}, FSS aims to utilize only a few labeled samples in the support set $\boldsymbol{S}$ to segment the objects in the query set $\boldsymbol{Q}$ and train the model in an episode manner to simulate few-shot scenarios~\cite{cao2023few,liu2022axial}. Specifically, each episode contains a support set and a query set with K-shot samples, denoted as:
\begin{enumerate}
\item[1)] a support set $\boldsymbol{S}=\left\{\left(\boldsymbol{I}_s^k, \boldsymbol{M}_{s, c}^k\right)\right\}_{c \in \boldsymbol{C}_{ {\mathrm{episode} }}}^{k=1, \ldots, K}$. Here, ${\boldsymbol{I}}_{{s}}^{{k}}$ denotes the $k$-th support image, while $\boldsymbol{M}_{{s,c}}^{{k}}$ represents the $k$-th mask for category $c$. Furthermore, $\boldsymbol{C}_{ {\mathrm{episode} }}$ signifies the category that aligns with the related episode.
\item[2)] a query set $\boldsymbol{Q}=\left\{\left(\boldsymbol{I}_q, \boldsymbol{M}_{q, c}\right)\right\}$, where $\boldsymbol{I}_q$ is the query image and $\boldsymbol{M}_{q, c}$ is the ground-truth mask for category $c$. Additionally, $\boldsymbol{M}_{{q}, {c}}$ is known during training and unknown in testing.
\end{enumerate}
There is only one defect category (one-way) in the support-query pair, and the other classes of defects will be regarded as background. 

\textbf{Local Descriptor-based Multi-Prototype}.
Advances in deep local descriptors have increasingly replaced traditional manual descriptors, primarily in few-shot classification applications \cite{li2019revisiting}. 
Yet, they are rarely used in FSS methods. 
We introduce a novel FSS strategy utilizing local descriptors in this paper. 
We begin by representing the feature map $\boldsymbol{F} \in \mathbb{R}^{c \times h \times w}$ as a local descriptor subset $\boldsymbol{X}=\left\{ \boldsymbol{x}_1, \boldsymbol{x}_2,..., \boldsymbol{x}_l\right\} \in \mathbb{R}^{c \times hw}$, where $c, h$ and $w$ are channel, height and width, respectively. And each local descriptor $\boldsymbol{x}$ corresponds to a 2-D representation vector in the feature map $\boldsymbol{F}$. In traditional metal surface defect FSS methods, defect features are represented as a prototype, typically embedded by support-query pairs. Our method refers to the subset of local descriptors containing defect information as multi-prototypes. This multi-prototype learning approach enables direct learning of subtler local cues, reducing the uncertainty inherent in traditional single prototype structures. We apply this method to tackle both intra-class semantic and distortion differences.

\textbf{Graph Convolution Network}.
Our network uniquely employs graph convolution on multi-prototypes based on local descriptors, differentiating it from other networks. Here, we define the graph structure ${\boldsymbol{\mathcal{G}}}=({V}, {E})$. In our model, $V$ signifies nodes and $E$ edges. We then define the leading matrix $\boldsymbol{A}$ and degree matrix $\boldsymbol{D}$, allowing us to express graph convolution as follows:
\begin{equation}
\label{EQ:gcn1}
    \begin{aligned}
       {\boldsymbol{H}^{(l+1)}}=\boldsymbol{\sigma} (\boldsymbol{\tilde{D}}^{-\frac{1}{2} } \boldsymbol{\tilde{A}}\boldsymbol{\tilde{D}}^{-\frac{1}{2} } \boldsymbol{H}^{(l)}\boldsymbol{\Theta}^{(l)} ),
    \end{aligned}
\end{equation}
where $\boldsymbol{H}^{(l+1)}$ means the feature of $(l+1)$-th layer, $\boldsymbol{H}^{(l)}$ means the feature of $l$-th layer, $\boldsymbol{\sigma}$ represents nonlinear activation function, $\boldsymbol{\tilde{A}}=\boldsymbol{I}+\boldsymbol{A}$, $\boldsymbol{\tilde{D}}_{ii}= {\textstyle \sum_{j}} \boldsymbol{\tilde{A}}_{ij}$, and $\boldsymbol{\Theta}^{(l)}$ represents the trainable parameters
corresponding to the features of the $l$-th layer. 
Next, We introduce graph Laplacian matrix to simplify the representation:
\begin{equation}
\label{EQ:gcn2}
    \begin{aligned}
       {\boldsymbol{\tilde{L}}}=\boldsymbol{\tilde{D}}^{-\frac{1}{2} } \boldsymbol{\tilde{A}}\boldsymbol{\tilde{D}}^{-\frac{1}{2} }.
    \end{aligned}
\end{equation}
Therefore, Eq. \eqref{EQ:gcn1} can be indicated to:
\begin{equation}
\label{EQ:gcn3}
    \begin{aligned}
       {\boldsymbol{H}^{(l+1)}}=\boldsymbol{\sigma} ({\boldsymbol{\tilde{L}}} \boldsymbol{H}^{(l)}\boldsymbol{\Theta}^{(l)} ).
    \end{aligned}
\end{equation}
Using these steps, we reason relationships between nodes in the graph structure for effective information extraction.

\textbf{Convolutional Block Attention Module}.
Convolutional block attention module (CBAM), used as auxiliary modules~\cite{feng2023detection,sheng2023faster}, activates target region features via partial convolution and activation functions to enhance network accuracy. In the channel attention module, we compress the channel by calculating the feature map's channel weight. This process is expressed as:
\begin{equation}
    \label{EQ:cSE}
    \begin{aligned}
        \boldsymbol{F}^{l+1}=\boldsymbol{F}^l \times \mathcal{Y} (\mathrm{avg}(\boldsymbol{F}^l)),
    \end{aligned}
\end{equation}
where $\mathcal{Y}(\cdot)$ represents nonlinear activation function and reshaping. $\mathrm{avg}(\cdot)$ denotes 2-D average pooling, $\boldsymbol{F}^{l+1}$ means the feature of $(l+1)$-th layer and $\boldsymbol{F}^l$ means the feature of $l$-th layer. Following the channel attention module, we conduct the spatial attention module, described as follows:
\begin{equation}
    \label{EQ:sSE}
    \begin{aligned}
        \boldsymbol{F}^{l+1}=\boldsymbol{F}^l \times \boldsymbol{\sigma} (\mathrm{Conv}(\boldsymbol{F}^l)),
    \end{aligned}
\end{equation}
where $\mathrm{Conv}(\cdot)$ means 2-D convolution. By CBAM, we effectively activate the features.

\section{Method}
In this section, we delineate the structure of LERENet, as illustrated in Figure~\ref{F:frame}. Initially, the approach utilizes local descriptors for the representation of image-level support and query features. These features are then individually embedded into graph and feature spaces, leading to the formation of local descriptor-based multi-prototype. Within the graph space, graph reasoning~\cite{chen2019graph,zhao2023dual} are applied to analyze defect relationships between support-query pair. Simultaneously, in the feature space, methodologies like masked average pooling and CBAM are implemented for the activation of defect features. Moreover, global edge information is harnessed to refine and guide the multi-prototype formation in the feature space, enhancing the efficacy of the provided information. The process culminates with the derivation of multi prototype $\boldsymbol{P}_\mathrm{main}$ from the graph space and multi prototype $\boldsymbol{P}_\mathrm{aux}$ from the feature space, followed by an information fusion process to yield the final segmentation output $\tilde{\boldsymbol{M}}_q$. For detailed algorithmic insights, please refer to the accompanying \textbf{Supplementary Materials}.



The network's data flow is succinctly described as follows: Local descriptors of the support-query pair are inputted into MPR and MPE, yielding $\boldsymbol{P}_{\mathrm{main}}$ and $\boldsymbol{P}_{\mathrm{aux}}$. Subsequently, $\boldsymbol{P}_{\mathrm{main}}$ and $\boldsymbol{P}_{\mathrm{aux}}$ are fused via IFM to derive the predicted mask $\boldsymbol{\tilde{M}}_q$. The loss of $\boldsymbol{\tilde{M}}_q$ and $\boldsymbol{M}_q$ are then computed utilizing a Binary Cross-Entropy (BCE) loss:
\begin{equation}
\label{EQ:loss}
    \begin{aligned}
       \boldsymbol{L}=\frac{1}{h w} \sum_{i=1}^h \sum_{j=1}^w \mathrm{BCE}\left(\boldsymbol{\tilde{M}}_q(i, j), \boldsymbol{M}_q(i, j)\right). 
    \end{aligned}
\end{equation}

\subsection{Multi-Prototype Reasoning}
Drawing from deep local descriptors and GCN~\cite{yao2024brain}, we embed local descriptors into graph space, encapsulating masked foreground information of support features and both foreground and background details of query features, termed as multi-prototypes. We then employ GCN to disseminate this information across adjacent prototypes, with a focus on the consistency in their distribution among local adjacent nodes. This learning stage provides the model with an initial comprehension of the defect areas in the query image. Given the query and masked support feature $\boldsymbol{F}_q, \boldsymbol{F}_s \in \mathbb{R}^{c \times h \times w}$, we utilize deep local descriptors to represent them as $\boldsymbol{X}_q, \boldsymbol{X}_s \in \mathbb{R}^{c \times hw}$. We analyze the consistency between defect features in support-query pairs and channels of each local descriptor. For this, we define multi-prototype $\boldsymbol{P}_{*}^v$ at the node level and multi-prototype $\boldsymbol{P}_{*}^d$ at the channel level, with intermediary $*$ representing either support or query feature. The process is represented as follows:
\begin{equation}
\label{EQ:pv}
\begin{aligned}
& \boldsymbol{P}_*^v=f_v\left(\boldsymbol{X}_*\right), 
\end{aligned} 
\end{equation}
\begin{equation}
\label{EQ:pd}
\begin{aligned}
& \boldsymbol{P}_*^d=f_d\left(\boldsymbol{X}_*\right),
\end{aligned} 
\end{equation}
where $\boldsymbol{X}_*$ means local descriptor-based query or support feature, $f_v(\cdot)$ and $f_d(\cdot)$ denote the 1-D convolution. Through Eqs.~\eqref{EQ:pv}-\eqref{EQ:pd}, we can obtain the multi-prototype at the node-level ${\boldsymbol{P}}_{{s}}^{{v}}, {\boldsymbol{P}}_{{q}}^{{v}}$ and channel level ${\boldsymbol{P}}_{{s}}^{{d}}, {\boldsymbol{P}}_{{q}}^{{d}}$.

We first define the relevance between the defect area of the support-query pair in graph space as $\boldsymbol{G}$. To achieve this, we further define relationship $\boldsymbol{G}_{sq}^v$ between multiple prototype nodes and relationship $\boldsymbol{G}_{sq}^d$ between channels of a single prototype. These two relationships can be obtained through the following steps:
\begin{equation}
\begin{aligned}
& \boldsymbol{G}_{sq}^{v}=\boldsymbol{P}_q^v \otimes \boldsymbol{P}_{s}^{d^T}, 
\end{aligned}
\label{EQ:g_sq_v}
\end{equation}
\begin{equation}
\begin{aligned}
& \boldsymbol{G}_{sq}^{d}=\boldsymbol{P}_q^d \otimes \boldsymbol{P}_{s}^{v^T}, 
\end{aligned}
\label{EQ:g_sq_d}
\end{equation}
where $\otimes$ represents matrix multiplication. We concatenate $\boldsymbol{G}_{sq}^v$ and $\boldsymbol{G}_{sq}^d$ along the channel direction to form $\boldsymbol{G}_{sq}$, representing the relationship between support and query defect features. Then, $\boldsymbol{G}$ is reasoned through GCN, with operations implemented by Eqs.(\ref{EQ:gcn1})-(\ref{EQ:gcn3}). This reasoning yields the relationship $\boldsymbol{G}$ in the graph space. To project this relationship onto defect features, we reflect it into the feature space using multi-prototype $\boldsymbol{P}_q^v$, obtaining $\boldsymbol{P}_\mathrm{main}$, which represents a multi-prototype annotated with supporting defect features. The entire process is represented as follows:
\begin{equation}
\begin{aligned}
\boldsymbol{P}_{ {\mathrm{main} }}=\boldsymbol{X}_q \oplus \boldsymbol{\tau}\left(\mathcal{R}(\boldsymbol{G} \otimes \boldsymbol{P}_q^v\right)),
\end{aligned}
\label{EQ:main}
\end{equation}
where $\oplus$ denotes element-wise addition, $\boldsymbol{\tau}(\cdot)$ represents 2-D convolution and normalization, and $\mathcal{R}(\cdot)$ reshapes the input to same size as $\boldsymbol{X}_q$. After that, we successfully obtain a multi-prototype $\boldsymbol{P}_{ {\mathrm{main} }} \in \mathbb{R}^{c \times hw}$ with defect information.

\begin{figure}[!htp]
\centering
\includegraphics[width=1\linewidth]{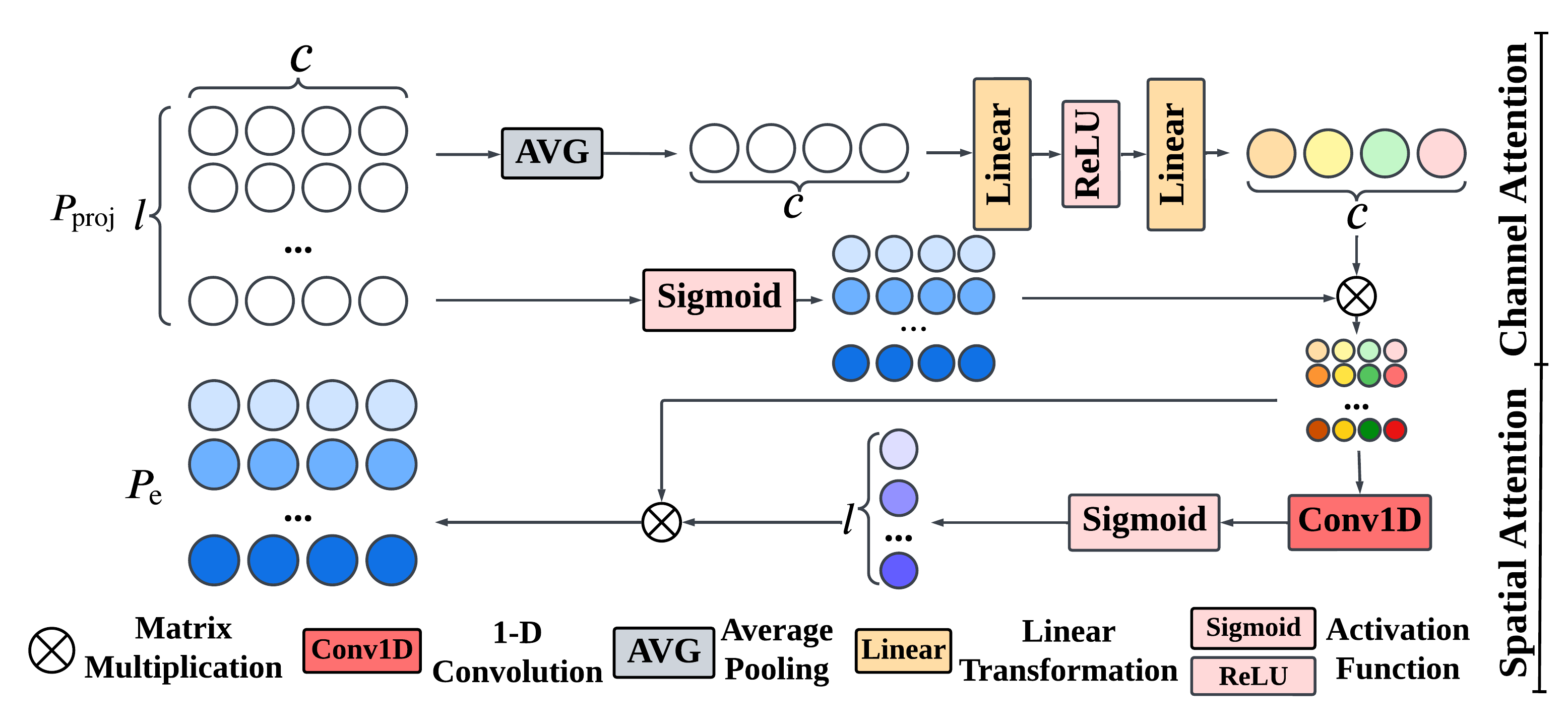}
\caption{The overall structure of CBAM. 
 It activates guided multi-prototype $\boldsymbol{P}_\mathrm{proj}$ via channel and spatial attention modules, and then yields the activated multi-prototype $\boldsymbol{P}_e$.}
\label{F:excitation}	
\end{figure}

\subsection{Multi-Prototype Excitation}
Despite distortion intra-class difference causing unclear local defect areas, the overall position of defects can still be approximated using supporting defect features. Drawing inspiration from masked average pooling~\cite{zhang2020sg} and spatial excitation in deep local descriptors~\cite{huang2021local}, we initially employ masked average pooling to pinpoint the approximate defect locations in the query image, then activate indistinct defect features using an incentive mechanism. This method allows us to identify key feature clues akin to human eye recognition. Specifically, firstly given the masked local descriptor-based support feature $\boldsymbol{X}_s$, we obtain a prototype of guidance information $\boldsymbol{X}_{ {\mathrm{proj} }} \in \mathbb{R}^{c \times 1}$ in the defect area through mask averaging pooling as:
\begin{equation}
    \begin{aligned}
        \boldsymbol{X}_{ {\mathrm{proj}}}=\mathrm{avg}\left(\boldsymbol{X}_s\right) \text {, }
    \end{aligned}
    \label{EQ:xproj}
\end{equation}
where $\operatorname{avg}(\cdot)$ represents 1-D average pooling with an output size of 1. Next, we embed guidance information $\boldsymbol{X}_{ {\mathrm{proj} }}$ from the support descriptors into the query descriptors to generate guided multi-prototype $\boldsymbol{P}_\mathrm{proj}$:
\begin{equation}
    \begin{aligned}
        \boldsymbol{P}_{ {\mathrm{proj} }}= \boldsymbol{X}_{\mathrm{proj}} \otimes \boldsymbol{X}_q  \text {, }
    \end{aligned}
    \label{EQ:pproj}
\end{equation}
As shown in Figure~\ref{F:excitation}, we obtain activated multi-prototype $\boldsymbol{P}_{\mathrm{e}}$ through CBAM by Eqs.~\eqref{EQ:cSE}-\eqref{EQ:sSE}, representing enhanced potential defect features for segmentation. Next, we reorganize this information into multi-prototype with complete defect features for downstream tasks, guided by global edge information. Specifically, we calculate similarity $\boldsymbol{D}_{\mathrm{cos}}$ between $\boldsymbol{f}_j \in \boldsymbol{F}_s$ and $\boldsymbol{f}_i \in \boldsymbol{F}_q$ by:
\begin{equation}
    \begin{aligned}
        \boldsymbol{D}_{\mathrm{cos}}=\frac{\boldsymbol{f}_i^T \boldsymbol{f}_j}{\left\|\boldsymbol{f}_i\right\| \| \boldsymbol{f}_j \|} \quad i, j \in\{1,2, \ldots, l\},
    \end{aligned}
    \label{EQ:s}
\end{equation}
where $\boldsymbol{D}_{\mathrm{cos}} \in \mathbb{R}^{{c} \times {c}}$. Next, we concatenate global information $\boldsymbol{D}_{\mathrm{cos}}$ with activated multi-prototype $\boldsymbol{P}_{\mathrm{e}}$, guiding the recombination of multi-prototypes with defect features to form multi-prototype $\boldsymbol{P}_\mathrm{aux} \in \mathbb{R}^{c \times l}$.

\subsection{Information Fusion Module}
Through upstream tasks, we have obtained multi-prototype $\boldsymbol{P}_\mathrm{main}$ from the graph space and multi-prototype $\boldsymbol{P}_\mathrm{aux}$ from the feature space. Multi-prototype $\boldsymbol{P}_\mathrm{main}$ contains defect feature information from supporting query pairs, while $\boldsymbol{P}_\mathrm{aux}$ holds finer clues and global edge information. We first integrate and share information between them, then extract it using two residual modules (Res), and ultimately derive segmentation mask $\tilde{\boldsymbol{M}}_q$ through the classification header (Cls). The entire process is outlined as follows:
\begin{equation}
    \begin{aligned}
        \boldsymbol{\tilde{\boldsymbol{M}}}_q=\mathcal{F}\left(\mathcal{R}\left(\mathcal{C}\left(\boldsymbol{P}_{ {\mathrm{main} }}, \boldsymbol{P}_{ {\mathrm{aux} }}\right)\right)\right),
    \end{aligned}
    \label{EQ:out}
\end{equation}
where $\mathcal{C}(\cdot)$ concatenates $\boldsymbol{P}_{ {\mathrm{main} }}$ and $\boldsymbol{P}_{ {\mathrm{aux} }}$, $\mathcal{R}(\cdot)$ reshapes the input to the same size as $\boldsymbol{M}_q$, and $\mathcal{F}(\cdot)$ indicates the Res and Cls.


\begin{table}[htbp]
  \centering
  \renewcommand{\arraystretch}{1.3}
  \scalebox{0.8}{
  \begin{tabular}{c|cccc}
\toprule[1pt]
Folds of Surface Defects-$4^{i}$ & \multicolumn{4}{c}{Categories}           \\ \hline
Surface Defects-$4^{0}$ & Al-I    & MT-I      & Steel-I & Steel-II \\
Surface Defects-$4^{1}$ & Leather & Steel-III & Rail    & Steel-IV \\
Surface Defects-$4^{2}$ & Al-II   & MT-II     & MT-III  & Tile     \\ \toprule[1pt]
\end{tabular}
  }
  \caption{Each flod of Surface Defects-$4^{i}$ dataset. Al-I and Al-II represent defects of rub mark and convexity on the aluminum surface, respectively. MT-I, MT-II, and MT-III represent defects of uneven, break, and fray on the magnetic tile. Steel-I, Steel-II, Steel-III, and Steel-IV represent defects of liquid, abrasion mask, patches, and scratches on the steel surface, respectively. In addition, leather and tile are two kinds of nonmetal data.}
  \label{tab:1}
\end{table}

\begin{table*}[ht]
  \centering
  
  \renewcommand{\arraystretch}{1.4}
  \scalebox{0.75}{
  \begin{tabular}{cccccccccccc}
\toprule[1pt]
\multicolumn{1}{c|}{\multirow{2}{*}{\textbf{Backbone}}} & \multicolumn{1}{c|}{\multirow{2}{*}{\textbf{Methods}}} & \multicolumn{5}{c|}{\textbf{1-shot}}                                                                       & \multicolumn{5}{c}{\textbf{5-shot}}                                                   \\
\multicolumn{1}{c|}{}                                   & \multicolumn{1}{c|}{}                                  & \textbf{Fold-0} & \textbf{Fold-1} & \textbf{Fold-2} & \textbf{mIoU}  & \multicolumn{1}{c|}{\textbf{FB-IoU}} & \textbf{Fold-0} & \textbf{Fold-1} & \textbf{Fold-2} & \textbf{mIoU}  & \textbf{FB-IoU} \\ \hline
\multicolumn{1}{c|}{\multirow{9}{*}{\textbf{VGG16}}}    & \multicolumn{1}{c|}{\textbf{ASGNet}~\cite{li2021adaptive}}                   & {\ul{35.65}}     & \textbf{41.65}  & {\ul{24.79}}     & {\ul{34.03}}    & \multicolumn{1}{c|}{{\ul{53.77}}}    & 38.50           & {\ul{34.23}}     & 25.86           & {\ul{32.86}}    & 51.16          \\
\multicolumn{1}{c|}{}                                   & \multicolumn{1}{c|}{\textbf{BAMNet}~\cite{lang2022learning}}                   & 11.58           & 18.05           & 16.77           & 15.47          & \multicolumn{1}{c|}{44.06}          & 27.76           & 27.19           & 26.49           & 27.15          & 46.88          \\
\multicolumn{1}{c|}{}                                   & \multicolumn{1}{c|}{\textbf{CANet}~\cite{zhang2019canet}}                    & 28.68           & 27.45           & 24.08           & 26.74          & \multicolumn{1}{c|}{52.22}          & 22.91           & 27.81           & 25.59           & 25.44          & 51.96          \\
\multicolumn{1}{c|}{}                                   & \multicolumn{1}{c|}{\textbf{DCPNet}~\cite{lang2022beyond}}                   & 31.10           & 28.91           & 21.26           & 27.09          & \multicolumn{1}{c|}{52.76}          & {\ul{47.33}}     & 20.00           & {\ul{27.97}}     & 31.77          & 51.45          \\
\multicolumn{1}{c|}{}                                   & \multicolumn{1}{c|}{\textbf{HDMNet}~\cite{peng2023hierarchical}}                   & 23.28           & 19.45           & 20.48           & 21.07          & \multicolumn{1}{c|}{51.14}          & 27.94           & 21.67           & 25.24           & 24.95          & {\ul{53.99}}    \\
\multicolumn{1}{c|}{}                                   & \multicolumn{1}{c|}{\textbf{PFENet}~\cite{tian2022prior}}                   & 16.98           & 14.85           & 13.22           & 15.02          & \multicolumn{1}{c|}{50.96}          & 17.11           & 15.23           & 13.66           & 15.33          & 51.22          \\
\multicolumn{1}{c|}{}                                   & \multicolumn{1}{c|}{\textbf{TGRNet(1-normal)}~\cite{bao2021triplet}}         & 29.78           & 25.15           & 24.36           & 26.43          & \multicolumn{1}{c|}{51.50}          & 37.42           & 24.66           & 26.52           & 29.53          & 53.27          \\
\multicolumn{1}{c|}{}                                   & \multicolumn{1}{c|}{\textbf{Baseline}}                 & 30.33           & 24.54           & 16.38           & 23.75          & \multicolumn{1}{c|}{49.12}          & 44.11           & 27.87           & 19.92           & 30.63          & 53.67          \\ \cline{2-12} 
\multicolumn{1}{c|}{}                                   & \multicolumn{1}{c|}{\textbf{LGRNet(Ours)}}             & \textbf{50.91}  & {\ul{33.72}}     & \textbf{25.11}  & \textbf{36.58} & \multicolumn{1}{c|}{\textbf{58.53}} & \textbf{49.63}  & \textbf{38.41}  & \textbf{28.16}  & \textbf{38.73} & \textbf{59.36} \\ 
\midrule
\midrule
\multicolumn{1}{c|}{\multirow{9}{*}{\textbf{ResNet50}}} & \multicolumn{1}{c|}{\textbf{ASGNet}~\cite{li2021adaptive}}                   & {\ul{36.49}}     & {\ul{40.71}}     & {\ul{28.00}}     & {\ul{35.07}}    & \multicolumn{1}{c|}{{\ul{56.67}}}    & 43.46           & 40.85           & 29.24           & {\ul{37.85}}    & 54.82          \\
\multicolumn{1}{c|}{}                                   & \multicolumn{1}{c|}{\textbf{BAMNet}~\cite{lang2022learning}}                   & 30.84           & 36.35           & 15.92           & 27.70          & \multicolumn{1}{c|}{53.38}          & 40.08           & 37.83           & 22.47           & 33.46          & 56.14          \\
\multicolumn{1}{c|}{}                                   & \multicolumn{1}{c|}{\textbf{CANet}~\cite{zhang2019canet}}                   & 26.22           & 20.69           & 12.56           & 19.82          & \multicolumn{1}{c|}{52.64}          & 28.85           & 22.98           & 14.40           & 22.08          & 52.78          \\
\multicolumn{1}{c|}{}                                   & \multicolumn{1}{c|}{\textbf{DCPNet}~\cite{lang2022beyond}}                   & 27.19           & 31.96           & 24.68           & 27.94          & \multicolumn{1}{c|}{51.67}          & {\ul{44.78}}     & 39.35           & {\ul{32.21}}     & 38.78          & {\ul{58.77}}    \\
\multicolumn{1}{c|}{}                                   & \multicolumn{1}{c|}{\textbf{HDMNet}~\cite{peng2023hierarchical}}                   & 35.58           & \textbf{40.79}  & 27.50           & 34.62          & \multicolumn{1}{c|}{56.01}          & 38.62           & {\ul{41.11}}     & \textbf{32.61}  & 37.45          & 56.19          \\
\multicolumn{1}{c|}{}                                   & \multicolumn{1}{c|}{\textbf{PFENet}~\cite{tian2022prior}}                   & 29.45           & 24.90           & 16.21           & 23.52          & \multicolumn{1}{c|}{54.06}          & 33.98           & 30.07           & 22.78           & 28.94          & 56.92          \\
\multicolumn{1}{c|}{}                                   & \multicolumn{1}{c|}{\textbf{TGRNet(1-normal)}~\cite{bao2021triplet}}         & 35.46           & 32.37           & 24.75           & 30.86          & \multicolumn{1}{c|}{53.62}          & 41.61           & 28.66           & 27.87           & 32.71          & 53.00          \\
\multicolumn{1}{c|}{}                                   & \multicolumn{1}{c|}{\textbf{Baseline}}                 & 34.84           & 24.17           & 19.63           & 26.21          & \multicolumn{1}{c|}{49.42}          & 43.55           & 24.53           & 23.98           & 30.69          & 50.66          \\ \cline{2-12} 
\multicolumn{1}{c|}{}                                   & \multicolumn{1}{c|}{\textbf{LGRNet(Ours)}}             & \textbf{51.89}  & 34.55           & \textbf{28.56}  & \textbf{38.33} & \multicolumn{1}{c|}{\textbf{58.77}} & \textbf{53.93}  & \textbf{41.23}  & 30.52           & \textbf{41.89} & \textbf{59.10} \\ \toprule[1pt]
\end{tabular}
  }
  \caption{Compare with state-of-the-art metal surface defect FSS and amateur networks on Surface Defect-$4^{i}$ in mIoU and FB-IoU under $1$-shot and $5$-shot. The \textbf{best} and \ul{second best} results are highlighted with \textbf{bold} and \ul{underline}.}
  
  \label{tab:2}%
\end{table*}%

\begin{table*}[htbp]
  \centering
  \renewcommand{\arraystretch}{1.1}
  \begin{tabular}{ccc|ccc|c|c}
\toprule[1pt]
\multirow{2}{*}{\textbf{MPR}} & \multirow{2}{*}{\textbf{MPE}} & \multirow{2}{*}{\textbf{MPE*}} & \multicolumn{3}{c|}{\textbf{1-shot/5-shot}}                        & \multirow{2}{*}{\textbf{mIoU}} & \multirow{2}{*}{\textbf{FB-IoU}} \\
                              &                               &                                & \textbf{Fold-0}      & \textbf{Fold-1}      & \textbf{Fold-2}      &                                &                                 \\ \hline
\Checkmark                             &                               &                                & 48.25/49.25          & 31.03/35.44          & 20.15/21.02          & 33.14/35.24                    & 57.96/55.83                     \\
                              & \Checkmark                             &                                & 28.29/32.51          & 20.25/21.04          & 23.42/25.51          & 23.99/26.35                    & 46.98/49.47                     \\
                              &                               & \Checkmark                              & 39.25/42.06          & 22.08/23.93          & 27.80/30.35          & 29.71/32.11                    & 52.54/56.23                     \\
\Checkmark                             & \Checkmark                             &                                & 49.64/50.64          & 33.04/35.92          & 21.71/28.03          & 34.80/38.19                    & 54.04/58.88                     \\
\Checkmark                             &                               & \Checkmark                              & \textbf{51.89/53.93} & \textbf{34.55/41.23} & \textbf{28.56/30.52} & \textbf{38.33/41.89}           & \textbf{58.77/59.10}            \\ \hline
\multicolumn{3}{c|}{\textbf{Baseline}}                                                         & 34.84/43.55          & 24.17/24.53          & 19.63/23.98          & 26.21/30.69                    & 49.42/50.62                     \\ \toprule[1pt]
\end{tabular}
  \caption{Ablation studies on each component. MPE* denotes MPE enhanced with global edge information.}
  \label{tab:ablation}%
\end{table*}

    

\section{Experiments}

\subsection{Experimental Settings}
\textbf{Dataset}.
To evaluate the effectiveness of LERENet, we report our results on the Surface Defect-$4^i$ dataset~\cite{bao2021triplet}, which comprises images and annotations of multiple metal surface defect dataset.
The total $12$ classes in this dataset are evenly
divided into $3$ folds, with each fold encompassing $4$ distinct classes.
As detailed in Table \ref{tab:1}, Surface Defect-$4^i$ (where $i\in{0,1,2}$) is our testing data in each fold, while the remaining part serves as our training data.

\textbf{Evaluation Metrics}. 
In alignment with existing FSS methods, we utilize the mean intersection-over-union (mIoU)~\cite{ding2023self,zheng2022quaternion} and the foreground-background IoU (FB-IoU)~\cite{zhao2023bmdenet,chen2022apanet} as our evaluation metrics.
While FB-IoU disregards object classes and calculates the average of foreground and background IoU directly, whereas mIoU calculates the average of the IoU values for all classes in a fold.

\textbf{Implementation Details}. 
We employ ResNet$50$~\cite{he2016deep} and VGG$16$~\cite{simonyan2014very} pretrained on Surface Defect-$4^i$ dataset as our backbone networks. 
At the same time, we utilize SGD optimizer to train $200$ epochs under the pytorch network framework.
For uniformity and consistency, the input image is resized to a standardized size of $200\times 200$.
Furthermore, we set the training batch size to $2$ and the test batch size to $1$, and conducted all the experiments with PyTorch1.12.1 and Nvidia GeForce RTX 3090 (20G) GPUs. 

\textbf{Baseline}. 
We remove the MPR, MPE and IFM from the LERENet to establish the baseline and retain only a Res module and a Cls module for downstream guidance.
The baseline leverages features extracted from the backbone, which are concatenated for input in downstream tasks to achieve final segmentation. 
The loss calculation in this approach is analogous to that in LERENet. 

\subsection{Comparison with State-of-the-Arts}
\textbf{Comparative Experiment}. 
We compare with several previous
state-of-the-art FSS algorithms on the Surface Defect-$4^i$ dataset. These algorithms encompass TGRNet which is specifically developed for metal surface defect segmentation, as well as six other general FSS algorithms. We report our results of Surface Defect-$4^i$ in Table \ref{tab:2}. It can be observed that LERENet is superior to other compared methods for mIoU and FB-IoU under both $1$-shot and $5$-shot settings. In the context of VGG$16$ backbone, our network demonstrates superior performance relative to TGRNet, evidencing an enhancement of $10.15 \%$ and $9.2\%$.  Meanwhile, LERENet demonstrates a substantial performance differential, achieving superiority by margins of $2.55 \%$ and $5.87 \%$ over the proximally ranked competitor. And we surpass PFENet, which is the top performer among six general FSS methods with VGG$16$ backbone, by $2.55 \%$ in $1$-shot setting and $5.87 \%$ in $5$-shot setting. Additionally, when we evaluated within the ResNet$50$ backbone, our performance is positioned the state-of-art. The mIoU of the baseline acts as a quantitative measure of the segmentation capacity provided by the backbone. The final segmentation efficacy of the model hinges on the advanced processing techniques employed subsequent to the baseline. Subsequently, we can find that its overall performance sometimes surpasses professional networks like VGG$16$ in $5$-shot scenarios. This is attributable to the fact that the baseline is adept at approximating query defect areas by leveraging supporting defect features. In contrast, randomness often characterizes the processing in other networks, potentially arising from random seeds or specific hyperparameters. It is noteworthy that occurrences, where the baseline mIoU exceeds performance in metal surface defect FSS, are incidental, predominantly manifesting within the VGG$16$ backbone. This contrast with ResNet$50$ suggests that the limited informational output of VGG$16$ may exacerbate the influence of adverse stochastic variables. ASGNet and HDMNet, notwithstanding their designation as amateur networks, demonstrate notable efficacy in addressing semantic intra-class variances. This effectiveness is principally ascribed to their sophisticated methodologies in pixel-level feature extraction and subsequent analytical processes. However, the proficiency of these networks in distortion intra-class difference is somewhat restricted. This is due to the imperative for feature enhancement. Overall, the accuracy of these networks sporadically exceeds ours. This trend, is largely attributable to their limitations on distortion intra-class difference, but frequently positions them as the second-most effective following our approach.

\begin{figure}[!h]
\centering
\includegraphics[width=0.9\linewidth]{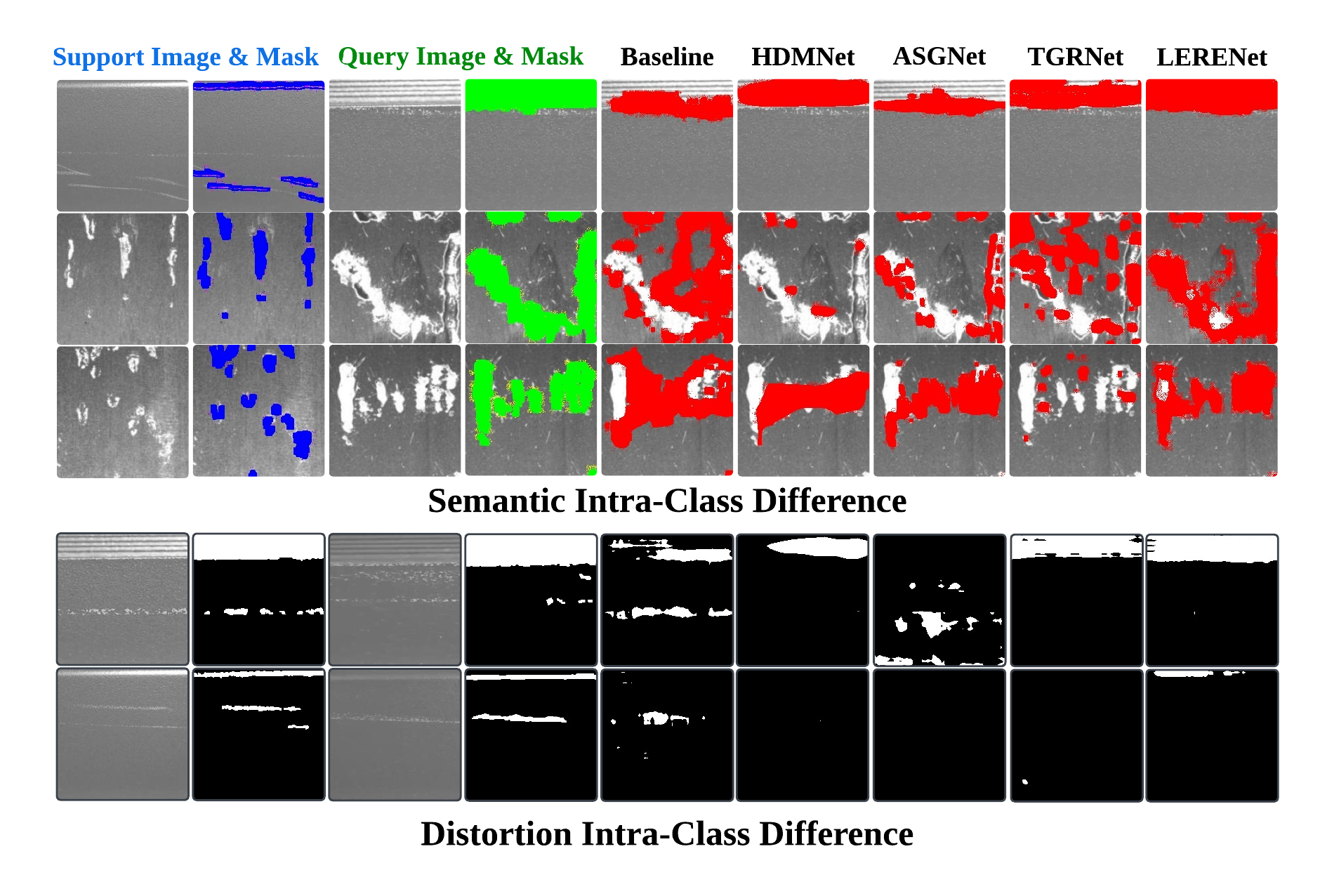}
\caption{Qualitative results of Baseline, HDMNet, ASGNet, TGRNet and our LERENet.}
\label{F:com_ex}	
\end{figure}

\textbf{Visualization}. 
Figure~\ref{F:com_ex} illustrates the selection of two distinct sets of support-query pairs, exemplifying semantic and distortion intra-class differences. 
Given the unique defect features, binary masks are employed to distinctly represent distortion intra-class difference. Concurrently, this figure also delineates the comparative performance of baseline, TGRNet, HDMNet and ASGNet, under the aegis of these two categories of differences, providing a visual exposition of their respective capabilities.
Our network consistently shows the most advanced effects for all different types. Additionally, it smoothly segments corresponding masks in cases of distortion intra-class differences, further validating our method's effectiveness.

\begin{figure}[!h]
\centering
\includegraphics[width=0.9\linewidth]{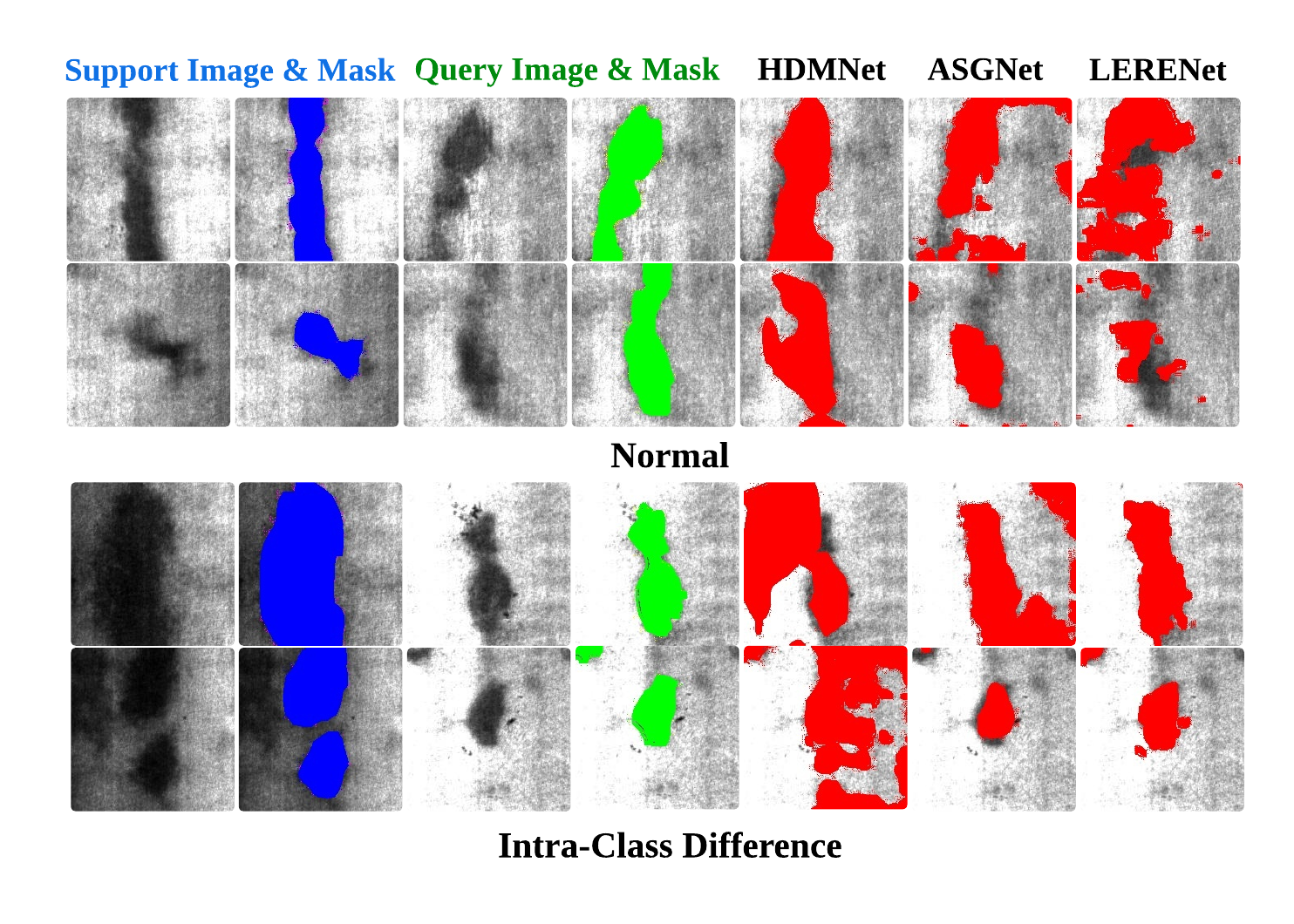}
\caption{Qualitative analysis of HDMNet, ASGNet, and our LERENet in patchy defects on steel surfaces.}
\label{F:amateur}	
\end{figure}

In Figure~\ref{F:amateur}, a selection of steel surface samples featuring plaque defects is employed to elucidate our specific limitations. 
This selection encompasses two distinct sample groups: one without notable intra-class differences and another exhibiting such differences. Analysis of these samples reveals that in scenarios devoid of significant intra-class difference, the MPE is prone to interference from background noise, which adversely affects segmentation efficacy. Conversely, in instances characterized by intra-class differences, the feature enhancement capabilities of MPE contribute to the superior performance of our network. For additional visualizations, consult the \textbf{Supplementary Materials}.

\subsection{Ablation Study}
We conduct ablation studies on MPR, MPE and MPE*, where MPE* denotes MPE guided by global edge information.
Table~\ref{tab:ablation} demonstrates that the individual incorporation of MPR and MPE* can result in an overall performance enhancement of $6.93 \%$ and $3.5 \%$, respectively, in comparison to the baseline.
The integration of these two modules results in an enhanced improvement of $12.12 \% $, further confirming the effectiveness and compatibility of our method.
In contrast, introducing MPE alone reduces mIoU below the baseline, but adding global edge information enhances it. 
This underscores the advantage of a motivated multi-prototype guided by global edge information. 
MPR significantly enhances semantic segmentation by analyzing the relevance between multi-prototype, showing the greatest overall contribution to our model, as evidenced by ablation studies. 
This confirms the superiority of our multi-prototype learning. 
However, MPR's impact is limited in Fold-2. Table~\ref{tab:ablation} analysis reveals the complementary relationship between MPR and MPE, indicating high compatibility without mutual interference. 
Specifically, MPR addresses semantic intra-class difference, while MPE tackles distortion intra-class difference, with their combination yielding substantial improvements. Furthermore, integrating global edge information into MPE significantly boosts its effectiveness. Thus, MPR's multi-prototype learning combined with MPE's multi-prototype motivation allows our network to excel in industrial scenarios.

\section{Conclusion}

This paper explores intra-class differences in metal surface defects, identifying semantic and distortion differences. To tackle these, we introduce \textbf{LERENet}, extracting features in support-and-query pairs by local descriptor-based multi-prototypes, integrated with MPR, and MPE to enhance the local and global view features to address two differences respectively. After that, we use IFM modules to fuse the information from the graph and feature space for semantic segmentation. Empirical tests confirm the effectiveness of our approach, highlighting its industrial applicability.


\newpage
\bibliographystyle{named}
\bibliography{ijcai24}

\newpage
\appendix
\section{The Training Process in \texttt{LERENet}.}
\label{sec:training}
The algorithm for the training process of LERENet is as follows:
\begin{algorithm}[htbp]
\DontPrintSemicolon
\caption{Local Descriptor-based Reasoning and Excitation Network (\texttt{LERENet})}
\SetAlgoLined 
\label{alg:LERENet}

\KwIn{A training dataset $\boldsymbol{D}_{\mathrm{train}}$}
\KwOut{Trained parameters}

\For{each episode $\left\{\left(\boldsymbol{I}_s^k, \boldsymbol{M}_{s, c}^k, \boldsymbol{I}_{q}, \boldsymbol{M}_{q,c}\right)\right\}_{c \in \boldsymbol{C}_{\mathrm{episode}}}^{k=1, \ldots, K} \in \boldsymbol{D}_{\mathrm{train}}$}{
    Extract $\boldsymbol{F}_{q}$ and masked $\boldsymbol{F}_{s}$ from backbone;
    
    Represent $\boldsymbol{F}_{s}$ and $\boldsymbol{F}_{q}$ as local descriptors $\boldsymbol{X}_{s}$ and $\boldsymbol{X}_{q}$;
    
    \tcp{Multi-Prototype Reasoning:}
    Embed to graph space by Eqs. \eqref{EQ:pv}-\eqref{EQ:pd};
    
    Obtain node and channel level relevance $\boldsymbol{G}_{sq}^{v}$ and $\boldsymbol{G}_{sq}^{d}$ by Eqs. \eqref{EQ:g_sq_v}-\eqref{EQ:g_sq_d};
    
    Concatenate $\boldsymbol{G}_{sq}^{v}$ and $\boldsymbol{G}_{sq}^{d}$ to get $\boldsymbol{G}_{sq}$;
    
    Reasoning $\boldsymbol{G}_{sq}$ through GCN to get $\boldsymbol{G}$ by Eqs. \eqref{EQ:gcn1}-\eqref{EQ:gcn3};
    
    Reflect back into feature space, and then get $\boldsymbol{P}_{\mathrm{main}}$ by Eq. \eqref{EQ:main};
    
    \tcp{Multi-Protorype Excitation:}
    Get support defect guidance $\boldsymbol{X}_{\mathrm{proj}}$ by Eq. \eqref{EQ:xproj};
    
    Generate multi-prototype $\boldsymbol{P}_{\mathrm{proj}}$ by Eq. \eqref{EQ:pproj};
    
    Calculate global edge information $\boldsymbol{D}_\mathrm{cos}$ by Eq. \eqref{EQ:s};
    
    Obtain multi-prototype $\boldsymbol{P}_{\mathrm{e}}$ by Eqs. \eqref{EQ:cSE}-\eqref{EQ:sSE};

    Concatenate $\boldsymbol{P}_{\mathrm{e}}$ and $\boldsymbol{D}_\mathrm{cos}$ to get multi-prototype $\boldsymbol{P}_{\mathrm{aux}}$;
    
    \tcp{Information Fusion Module:}
    Fuse $\boldsymbol{P}_{\mathrm{main}}$ and $\boldsymbol{P}_{\mathrm{aux}}$ by Eq. \eqref{EQ:out} to get $\boldsymbol{\tilde{M}}_{q}$;
    
    Compute binary cross entropy Loss $\boldsymbol{L}$ by Eq. \eqref{EQ:loss};
}

    
    
    

    

\end{algorithm}

In the above algorithm, $\boldsymbol{F}_s$ and $\boldsymbol{F}_q$ signify image-level features, while $\boldsymbol{X}_s$, $\boldsymbol{X}_q$ and $\boldsymbol{X}_\mathrm{proj}$ denote local descriptors-based features. $\boldsymbol{G}_{sq}^v$, $\boldsymbol{G}_{sq}^d$, $\boldsymbol{G}_{sq}$ and $\boldsymbol{G}$ illustrate the relationships between support-query pairs at varying levels within the graph space. The multi-prototype $\boldsymbol{P}_\mathrm{main}$ is the output of the multi-protorype reasoning (MPE). $\boldsymbol{P}_\mathrm{proj}$ embodies the guided multi-prototype, and $\boldsymbol{P}_e$ represents the multi-prototype following convolutional block attention module (CBAM). $\boldsymbol{D}_\mathrm{cos}$ encapsulates global edge information, and the multi-prototype $\boldsymbol{P}_\mathrm{aux}$ signifies the output from the multi-prototype excitation (MPE). The testing process mirrors the training process in its entirety, with the sole exception being that its output is only a mask.

\section{Experiment}
\subsection{Implement details}
\begin{enumerate}
    \item [1)] In extension to K-shot (K $>$ 1) setting, K support
images with their annotated masks $\boldsymbol{S}=\left\{\left(\boldsymbol{I}_s^k, \boldsymbol{M}_{s, c}^k\right)\right\}_{c \in \boldsymbol{C}_{ {\mathrm{episode} }}}^{k=1, \ldots, K}$ and the query set $\boldsymbol{Q}=\left\{\left(\boldsymbol{I}_q, \boldsymbol{M}_{q, c}\right)\right\}$ are given. LERENet can be quickly and easily extended to the new setting based on the feature averaging. Subsequently, we derive the masked $\boldsymbol{F}$ by:
\begin{equation}
    \begin{aligned}
        \boldsymbol{F}_{s}=\frac{ {\textstyle \sum_{i=1}^{K}}\boldsymbol{\varphi}(\boldsymbol{I}_s^i)  }{K} ,
    \end{aligned}
    \label{EQ:k}
\end{equation}
where $\boldsymbol{\varphi}(\cdot)$ denotes feature extraction. By doing so, we integrating information from multiple shots streamlines subsequent operations.
    \item [2)] In the multi-prototype reasoning (MPR) and multi-protorype excitation (MPE), the middle-level features are obtained by backbone. For instance, we get the middle-level features of ResNet$50$ through concatenating the features from block 2 and block 3. The middle-level feature dimension $c$ is 256.
\end{enumerate}

\subsection{Comparison with State-of-the-art Methods}

\begin{table}[!h]
\centering
\scalebox{0.65}{
\begin{tabular}{c|cccc|c}
\hline
\multirow{2}{*}{\textbf{Methods}} & \multicolumn{4}{c|}{\textbf{mIoU}} & \multirow{2}{*}{\textbf{Mean FB-IoU}} \\
 & \textbf{Fold-0} & \textbf{Fold-1} & \textbf{Fold-2} & \textbf{Mean} &  \\ \hline
\textbf{ASGNet~\cite{li2021adaptive}} & 35.93 & 34.87 & 19.92 & 30.24 & 52.39 \\
\textbf{HDMNet~\cite{peng2023hierarchical}} & 41.65 & 35.73 & \textbf{29.84} & 35.74 & 52.85 \\
\textbf{TGRNet~\cite{bao2021triplet}} & 36.40 & 37.96 & 24.33 & 32.90 & 52.46 \\
\textbf{LERENet} & \textbf{45.88} & \textbf{38.56} & 25.87 & \textbf{36.77} & \textbf{53.48} \\ \hline
\end{tabular}
}
\caption{Performance comparison on Surface Defect-$4^i$ when using ResNet$101$.}
\label{tab:ex1}%
\end{table}

\begin{table}[!h]
\centering
\scalebox{1}{
\begin{tabular}{c|cccc}
\hline
\textbf{Methods} & \textbf{mIoU} & \textbf{FB-IoU} & \textbf{FLOPs} & \textbf{\#Params.} \\ \hline
\textbf{TGRNet} & 30.86 & 53.62 & 83.69G & 9.38M \\
\textbf{LERENet} & \textbf{38.33} & \textbf{58.77} & \textbf{69.85G} & \textbf{8.28M} \\ \hline
\end{tabular}
}
\caption{Comparison of model performance. ``FLOPs" indicates
the computational overhead. ``\#Params." indicates the number of learnable parameters.}
\label{tab:exFlops}%
\end{table}

Initially, we delineate the outcomes derived from employing ResNet$101$ as the backbone under 1-shot settings, as systematically detailed in Table~\ref{tab:ex1}. It can be seen that our approach achieves new state-of-the-art performance and outperforms previous state-of-the-art result by $1.03 \%$ and $0.63\%$. Subsequently, with an emphasis on cost efficiency in industrial contexts, we compared FLOPs and parameters in Table~\ref{tab:exFlops}. This comparison reveals that our method exhibits lower FLOPs and fewer parameters, facilitating swifter completion of the corresponding defect segmentation tasks.

\subsection{Ablation Study}
We executed supplementary ablation studies to rigorously assess the influence of our designs. Note that the experiments in this section are performed on Surface Defect-$4^i$ dataset using the ResNet$50$ backbone unless specified otherwise. Furthermore, the evaluation metric is the mIoU and FB-IoU.
\begin{table}[!h]
\centering
\scalebox{0.85}{
\begin{tabular}{c|cccc|c}
\hline
\multirow{2}{*}{\textbf{Pretrained}} & \multicolumn{4}{c|}{\textbf{mIoU}} & \multirow{2}{*}{\textbf{Mean FB-IoU}} \\
 & \textbf{Fold-0} & \textbf{Fold-1} & \textbf{Fold-2} & \textbf{Mean} &  \\ \hline
 & 46.99 & 33.25 & 26.69 & 35.64 & 53.30 \\
\checkmark & \textbf{51.89} & \textbf{34.55} & \textbf{28.56} & \textbf{38.33} & \textbf{58.77} \\ \hline
\end{tabular}
}
\caption{Ablation studies of the pretrained strategy.}
\label{tab:pre}%
\end{table}

\textbf{Effect of the pretrained strategy}.
Current state-of-the-art methods usually incorporate pretrained strategy prior to engaging in meta-learning~\cite{cao2024few,10382511}, with the objective of enhancing overall performance. The  experiment presented in Table~\ref{tab:pre} provides a robust demonstration of the efficacy of this strategy.

\begin{table}[!h]
\centering
\scalebox{0.85}{
\begin{tabular}{c|cccc|c}
\hline
\multirow{2}{*}{\textbf{Loss}} & \multicolumn{4}{c|}{\textbf{mIoU}} & \multirow{2}{*}{\textbf{Mean FB-IoU}} \\
 & \textbf{Fold-0} & \textbf{Fold-1} & \textbf{Fold-2} & \textbf{Mean} &  \\ \hline
\textbf{Focal} & 46.35 & 34.12 & 21.23 & 33.90 & 55.59 \\
\textbf{Dice} & 33.46 & 14.57 & 22.74 & 23.59 & 40.43 \\
\textbf{BCE+Dice} & 43.56 & \textbf{38.90} & 24.96 & 35.81 & 56.35 \\
\textbf{BCE} & \textbf{51.89} & 34.55 & \textbf{28.56} & \textbf{38.33} & \textbf{58.77} \\ \hline
\end{tabular}
}
\caption{Ablation studies of the loss strategy}
\label{tab:exloss}%
\end{table}

\textbf{Effect of the loss strategy}.
Recently, many different losses have been proposed to replace cross-entropy loss in the field of segmentation. They are usually used to alleviate the problem of imbalance between positive and negative samples in the foreground and background (such as Focal loss). Therefore, ablation studies in Table~\ref{tab:exloss} were conducted to determine an optimal loss for LERENet. It is evident from our analysis that the BCE loss demonstrates a more facile optimization process compared to other loss functions. Consequently, we employ the BCE loss in our network.
\subsection{More Visualizations}
\begin{figure*}[!t]
\centering
\includegraphics[width=0.8\linewidth]{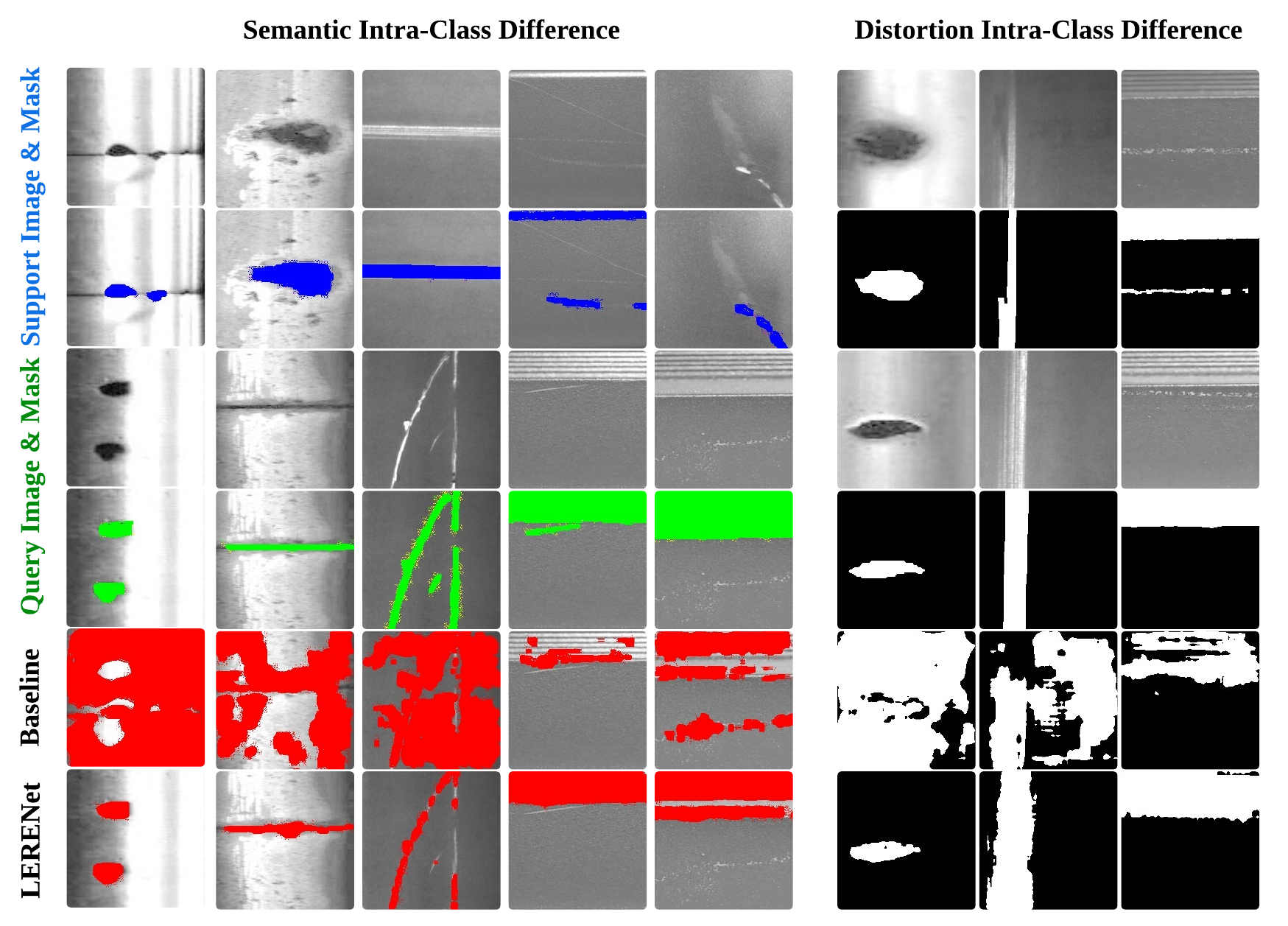} 
 \caption{Qualitative results of our method LERENet and baseline on Surface Defect-$4^i$.}
\label{F:extra}	
\end{figure*}

\begin{figure*}[!t]
\centering
\includegraphics[width=0.8\linewidth]{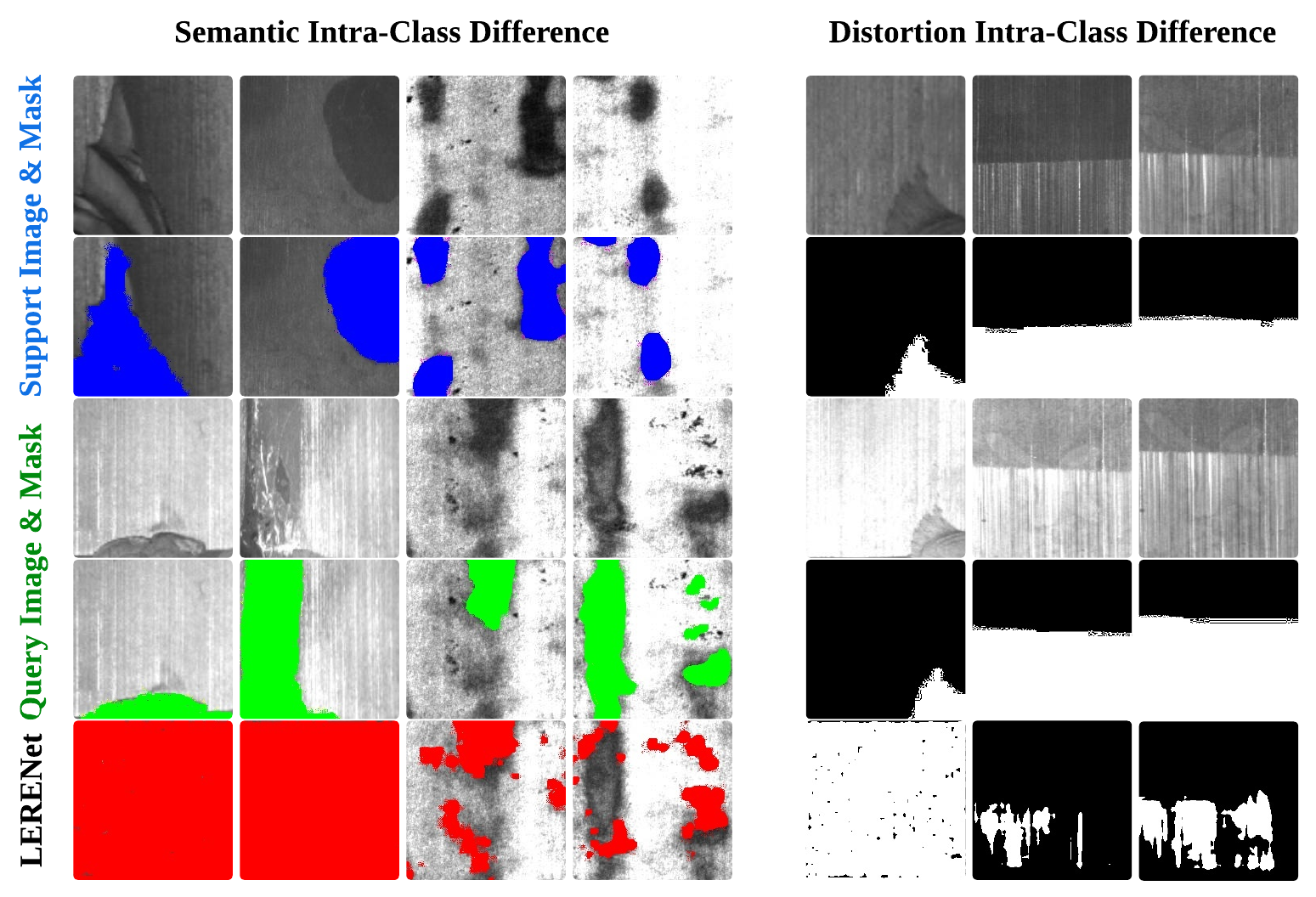} 
 \caption{Failure results of our method LERENet on Surface Defect-$4^i$.}
\label{F:fail}	
\end{figure*}

In Figure~\ref{F:extra}, we showcase additional qualitative results, demonstrating that our network adeptly handles in semantic and distortion intra-class differences. This highlights the refined ability if network in discerning and addressing the intricate intra-class difference typical in such defects. Moreover, some failure cases are also provided in Figure~\ref{F:fail}. As the Figure~\ref{F:fail} shows, we can conclude that (1) Samples exhibiting both semantic and distortion intra-class differences significantly impede segmentation accuracy (1st and 2nd columns). (2) Sample noise detrimentally affects segmentation precision (3rd and 4th columns). (3) In the context of perspective distortion, intra-class difference markedly compromise segmentation efficacy (5th, 6th, and 7th columns). 

\label{sec:visualization}

\end{document}